# Separating a Real-Life Nonlinear Image Mixture

**Luís B. Almeida**
*Instituto de Telecomunicações, Instituto Superior Técnico*
*Av. Rovisco Pais, 1*
*1049-00 Lisboa, Portugal*
*Phone: +351-218417679*
*Fax: +351-218418472*                                    LUIS.ALMEIDA@LX.IT.PT

**Editor:**

## Abstract

When acquiring an image of a paper document, the image printed on the back page sometimes shows through. The mixture of the front- and back-page images thus obtained is markedly nonlinear, and thus constitutes a good real-life test case for nonlinear blind source separation.

This paper addresses a difficult version of this problem, corresponding to the use of "onion skin" paper, which results in a relatively strong nonlinearity of the mixture, which becomes close to singular in the lighter regions of the images. The separation is achieved through the MISEP technique, which is an extension of the well known INFOMAX method. The separation results are assessed with objective quality measures. They show an improvement over the results obtained with linear separation, but have room for further improvement.

## 1. Introduction

When an image of a paper document is acquired, e.g. through scanning, photographing or photocopying, the image printed on the back page sometimes shows through. This is normally due to partial transparency of the paper, and results in the acquisition of a mixture of the images from the front and back pages. It is usually possible to obtain two different mixtures, by acquiring both sides of the document. This is a situation that seems suited for handling by blind source separation (BSS) techniques. The main difficulty is that the images that are acquired are nonlinear mixtures of the original images printed on each of the sides of the paper. This is, therefore, an interesting test case for nonlinear BSS methods, with potential application in scanners, photocopiers and in document processing in general.

This paper addresses a difficult instance of this problem, in which the paper that is used is of the "onion skin" type. This creates a mixture that has a relatively strong nonlinearity, and that is close to singular in the lighter parts of the images. For separation we use MISEP, which is a nonlinear independent component analysis (ICA) technique (Almeida, 2003b). MISEP is a generalization of the well known INFOMAX technique of linear ICA (Bell and Sejnowski, 1995), extending it in two directions: (1) being able to handle nonlinear mixtures, and (2) using output nonlinearities that adapt to the statistical distributions of the extracted components.

Besides the separation itself, an important practical issue in this specific situation is the alignment of the two mixture images. One might think that, by an appropriate translation and rotation, the images from the two sides of the document could be brought into good alignment with each other. It was found, however, that scanners normally introduce slight geometrical distortions that make it necessary to use local alignment techniques to obtain an image alignment that is adequate for separation. That alignment issue is also addressed in this paper, because it is an important step of the image processing that needs to be done.

Published results concerning nonlinear BSS in real-life problems are still very few. To the author's knowledge, and apart from an earlier version of the present work (Almeida and Faria, 2004), the only published report of blind source separation of a real-life nonlinear mixture in which the recovery of the original sources can be confirmed is (Haritopoulos et al., 2002). Some other applications of nonlinear ICA to real-life data, e.g. (Lappalainen and Honkela, 2000), (Lee and Batzoglou, 2003), don't provide means to confirm whether real sources were recovered.

This manuscript's structure is as follows: Section 2 provides a brief overview of non-linear separation methods. Section 3 presents a short summary of the MISEP method, to outline its basic principles and to set the notation. Section 4 describes the experimental conditions, including image printing, acquisition and alignment. Section 5 presents the experimental results, which are assessed with objective measures of separation quality. Section 6 concludes.

In the printed version of this paper some of the details of some images may be lost due to the printing process. However, the paper is freely available online, and in the electronic online version one can zoom in on the images (scatter plots and images of sources, mixtures and separated components) to better view the details. In the pdf version ($\sim 7$ MB) the images are encoded in JPEG format and therefore show some artifacts, which become noticeable on close inspection. The postscript version shows the images without artifacts, but corresponds to a larger file ($\sim 14$ MB). The two versions are available at

http://www.lx.it.pt/~lbalmeida/papers/AlmeidaJMLR05.pdf
http://www.lx.it.pt/~lbalmeida/papers/AlmeidaJMLR05.ps.zip

The source and mixture images used in this paper are available online at
http://www.lx.it.pt/~lbalmeida/ica/seethrough.
The separation routines that were used to produce the results are available at
http://www.lx.it.pt/~lbalmeida/ica/seethrough/code/jmlr05.

## 2. Overview of nonlinear ICA methods

In this Section we provide a short overview of some of the main nonlinear ICA methods. This overview is necessarily very brief, and the reader is referred to an overview paper (Jutten and Karhunen, 2004) for more complete information.

It is interesting to note that one of the very early works on ICA (Schmidhuber, 1992) already proposed a nonlinear method. Although being based on an interesting principle (minimization of predictability of each extracted component by the other components) it was rather unpractical and computationally heavy.

The essential uniqueness of the solution of linear ICA (Comon, 1994), together with the greater simplicity of linear separation and with the fact that many naturally occurring mixtures are essentially linear, led to a quick development of linear ICA. The work on nonlinear ICA probably was slowed down mostly by its inherent ill-posedness and by its greater complexity, but development of nonlinear methods has continued steadily (e.g. Burel, 1992; Deco and Brauer, 1995; Marques and Almeida, 1999; Palmieri et al., 1999; Theis et al., 2003). The methods that have received the strongest attention in recent years are very briefly outlined in the next paragraphs.

Ensemble learning (Lappalainen and Honkela, 2000) is a Bayesian method and, as such, uses prior distributions as a form of regularization, to handle the ill-posedness problem. It is computationally heavy, but has produced some interesting results, including an extension to the separation of nonlinearly mixed dynamical processes (Valpola and Karhunen, 2002).

Kernel-based nonlinear ICA (Harmeling et al., 2003) essentially consists of linear ICA performed on a high-dimensional space that is a nonlinear transformation of the original space of mixture observations. In the form in which it was presented in the cited reference, it used the temporal structure of the signals to perform the linear ICA operation. This apparently helped it to effectively deal with the ill-posedness problem, and allowed it to yield some impressive results on artificial, strongly nonlinear mixtures. The method seems to be quite tractable, in computational terms.

MISEP (Almeida, 2003b) is an extension of INFOMAX (Bell and Sejnowski, 1995) into the nonlinear domain. It uses regularization to deal with the ill-posedness problem, and is computationally tractable. It is described in more detail in the next Section, since it is the method used in the present paper.

A special class of methods that deserves mention deals with nonlinear mixtures which are constrained so as to make the result of ICA essentially unique, as in linear ICA. The most representative class corresponds to the so-called post-nonlinear (PNL) mixtures (Taleb and Jutten, 1999). These are linear mixtures followed by component-wise invertible non-linearities. The interest of this class resides both in its unique separability and in the fact that it corresponds to well identified practical situations: linear mixtures observed by non-linear sensors. PNL mixtures and their extensions have had a considerable development (see Jutten and Karhunen (2004) for references).

## 3. Overview of the MISEP method

MISEP (Almeida, 2003b) is a generalization of the INFOMAX method of linear ICA (Bell and Sejnowski, 1995). We recall that the latter method, although initially introduced under a principle of maximum information preservation, was later shown to be interpretable as a maximum likelihood method (Pearlmutter and Parra, 1997), and also as a method based on the minimization of the mutual information (MI) of the extracted components (Hyvärinen and Oja, 2000). We briefly recall the latter interpretation, albeit using a reasoning different from the one given in that reference.

If $\mathbf{Y}$ is a vector with random components $Y_i$, we define the mutual information of the components of $\mathbf{Y}$ as

$$I(\mathbf{Y}) = \sum_i H(Y_i) - H(\mathbf{Y}) \tag{1}$$

where, for continuous variables, as is the case here, $H$ denotes Shannon's differential entropy

$$H(X) = -\int p(x) \log p(x) \mathrm{d}x. \tag{2}$$

In this equation $p(x)$ is the probability density of the scalar random variable $X$ (we denote probability density functions by $p(\cdot)$, the function's argument clarifying which random variable is being considered; this is a slight abuse of notation, but helps to keep expressions simpler and does not create any confusion). A similar definition holds for $H(\mathbf{X})$, where $\mathbf{X}$ is a random vector, the difference being that the random variable is now multidimensional and the integral in (2) becomes a multiple integral, encompassing the whole domain of $\mathbf{X}$.

Mutual information is a good measure of statistical dependence. $I(\mathbf{Y})$ measures the amount of information that is shared among the random variables $Y_i$. It is always positive, except if these variables are mutually statistically independent, in which case it is zero. $I(\mathbf{Y})$ is also equal to the Kullback-Leibler divergence between the product of the marginal densities, $\prod_i p(y_i)$ and the true joint density, $p(\mathbf{y})$. These two densities are equal if and only if the components $Y_i$ are mutually independent.

Minimization of the mutual information of the extracted components is therefore a good criterion for independent component analysis. An interesting and useful property of mutual information, that we shall use ahead, is that if we apply invertible, possibly nonlinear, transformations to the random variables, $Z_i = \psi_i(Y_i)$, the mutual information doesn't change: $I(\mathbf{Z}) = I(\mathbf{Y})$.

INFOMAX uses a network with the structure depicted in Fig. 1. Block $\mathbf{F}$ performs the separation proper, the separated components being $y_i$. $\mathbf{F}$ is linear, corresponding just to a product by a matrix. The blocks $\psi_i$ are auxiliary, being used only during the training phase. Each of these blocks performs an invertible, increasing transformation $z_i = \psi_i(y_i)$, whose counter-domain is the interval $[0, 1]$.

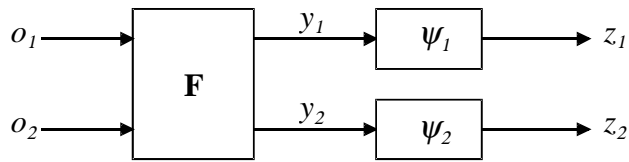

Figure 1: Network structure used in INFOMAX and in MISEP. In INFOMAX, $\mathbf{F}$ is an adaptive linear block, and the $\psi_i$ are fixed a priori. In MISEP, $\mathbf{F}$ can be nonlinear, and both $\mathbf{F}$ and $\psi_i$ are adaptive.

If we choose each $\psi_i$ as the cumulative distribution function (CDF) of the corresponding $Y_i$, it is easy to see that each of the $Z_i$ will be uniformly distributed in $[0, 1]$, resulting in

$p(z_i) = 1$ for $z_i$ in that interval, and $H(Z_i) = 0$. Therefore,

$$I(\mathbf{Y}) = I(\mathbf{Z}) \tag{3}$$

$$= \sum_i H(Z_i) - H(\mathbf{Z}) \tag{4}$$

$$= -H(\mathbf{Z}) \tag{5}$$

Mutual information is hard to minimize directly, but (5) shows that, under the stated conditions, this minimization is equivalent to the maximization of the output entropy $H(\mathbf{Z})$, a maximization which is much easier to achieve. INFOMAX works by optimizing $\mathbf{F}$ such that $H(\mathbf{Z})$ is maximized. We won't go into the details here, but the reader can consult (Bell and Sejnowski, 1995) or (Hyvärinen and Oja, 2000) for a deeper discussion.

As said above, MISEP extends INFOMAX in two directions. The first is being able to deal with nonlinear mixtures. This is achieved by allowing block $\mathbf{F}$, in Fig. 1, to be nonlinear. We have often implemented this block by means of a multilayer perceptron (MLP), but essentially any adaptive nonlinear structure can be used. For example, a radial basis function network has been used in Almeida (2003a), and a specialized structure in Almeida and Faria (2004).

The second direction in which MISEP extends INFOMAX, is by making the output transformations $\psi_i$ adaptive. As we have seen above, each $\psi_i$ should correspond to the CDF of the corresponding extracted source, for the maximization of the output entropy to correspond to the minimization of the mutual information of the extracted components. The a-priori choice of the $\psi_i$ functions in INFOMAX can be seen as a user-made, prior assumption about the distributions of the sources. In MISEP the $psi_i$ blocks are adaptive, being implemented by means of adequately constrained MLPs. It can be shown that maximization of the output entropy $H(\mathbf{Z})$ leads each of these blocks to estimate the corresponding CDF, while simultaneously leading $\mathbf{F}$ to minimize the mutual information $I(\mathbf{Y})$ (Almeida, 2003b). Therefore, maximizing the output entropy simultaneously adapts the $\psi_i$ blocks and leads to the minimization of the mutual information $I(\mathbf{Y})$.

An issue that has frequently been discussed is whether nonlinear blind source separation, based on ICA, is feasible in practice. This debate has to do with the fact that nonlinear ICA, with no additional constraints, is an ill-posed problem, having an infinite number of solutions that are not related to one another in any simple way (Darmois, 1953; Hyvärinen and Pajunen, 1999; Marques and Almeida, 1999). Therefore we cannot expect that, just by extracting independent components, one will be able to recover the original sources that were nonlinearly mixed. This is to be contrasted with the situation in linear ICA/BSS in which, under very mild constraints, there exists essentially only one solution (Comon, 1994). In linear ICA, if independent components are extracted, they must correspond to the original sources, apart from possible scaling and permutation. This author has argued that in the nonlinear case, when the mixture is not too strongly nonlinear, adequate regularization should allow the handling of the ill-posedness of nonlinear ICA, still allowing the approximate recovery of the sources. The nonlinearities considered in this paper would be classified by the author as of "medium intensity". As we shall see below, approximate source recovery was possible, and the indetermination of nonlinear ICA didn't lead to inadequate separation.

## 4. Experimental setup

In this section we describe the experimental setup, including details of image printing, acquisition and preprocessing

### 4.1 Source images

We used five image mixtures as test cases. The corresponding pairs of source images are shown in Figs. 2 and 3. The main properties of these image pairs are as follows:

1. In the first pair, each image consists of 25 uniform bars with intensities that are uniformly spaced between black and white, and are randomly ordered. The first image has vertical bars, and the second image is just the first one rotated by $90^o$. Thus, by construction, the intensities of the two images are independent, and each of the images has an intensity distribution which is close to uniform.

2. The second pair consists of images of natural scenes with a relatively high degree of variability and relatively small details. This causes a strong "mixing" of intensities, and the two sources are approximately independent from each other. However, the small details tend to make image superposition (due to imperfect separation) hard to notice visually.

3. The third pair consists of an image of a natural scene, on one side of the paper, and an image of printed text (Times New Roman, 12-point font) on the other side. Since the text has many large changes of intensity in very small areas, a good "mixing" of the intensities from both images takes place, and the two sources are approximately independent.

4. The fourth pair consists of printed text on both sides of the paper, with a few graphs on one of the sides. Once again, the intensities from the two sides of the paper are well mixed, and therefore approximately independent. The peculiarity of this pair is that, since printed text has a much larger area of white than of black, only a very small percentage of pixels is simultaneously dark on both sides of the paper. This has some influence on the separation results that are obtained, as we shall see.

5. The fifth pair consists of images of natural scenes that have large areas with quasi-uniform intensity. This causes a relatively weak mixing of intensities, making the intensities from the two sides of the paper non-independent. This fact has some impact on the separation results, as we shall see. The large, relatively uniform areas of the images make imperfect separation easier to notice visually than in case 2 above.

The leftmost columns of Figs. 4 and 5 illustrate the joint distributions of the source images. These plots deserve some comments. First of all we should note that, for the joint distributions of the two sources of each pair to be meaningful, the source images had to be adjusted in resolution and aligned, so as to be in the same relative position as in the acquired mixtures. For that purpose each source image was reduced in resolution to the same size as the corresponding acquired mixture images, and was then aligned with the corresponding separated component from nonlinear separation (see Section 4.3 for the alignment procedure and Section 5.2 for the nonlinear separation procedure). Both the resizing and the alignment procedures involved bicubic interpolation of the pixel intensities. The result of such interpolation is visible in the edges of the bars and of the text characters, in Figs. 2 and 3, which show the source images after resizing and alignment.

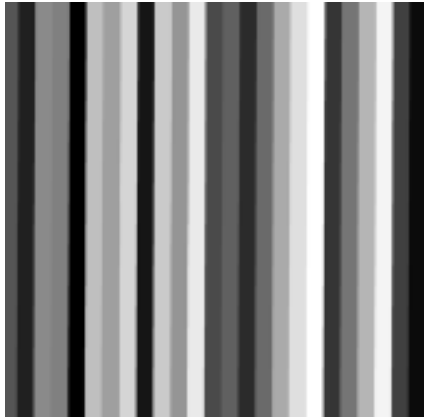 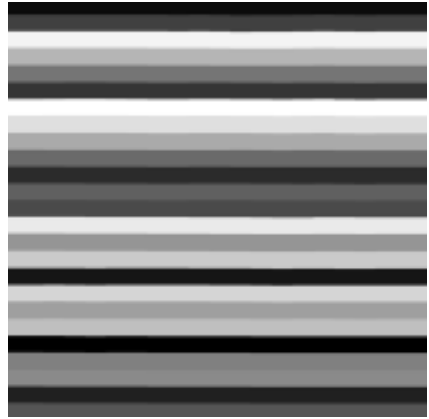

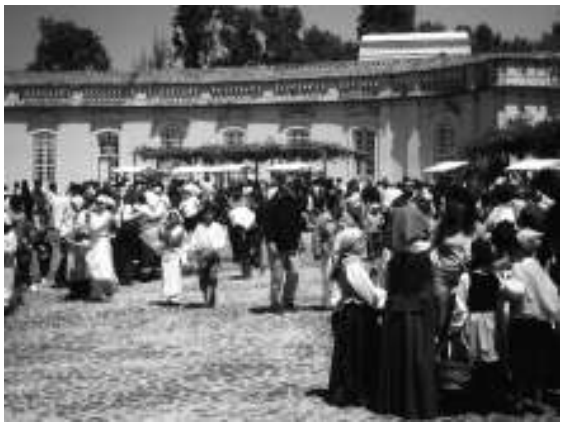 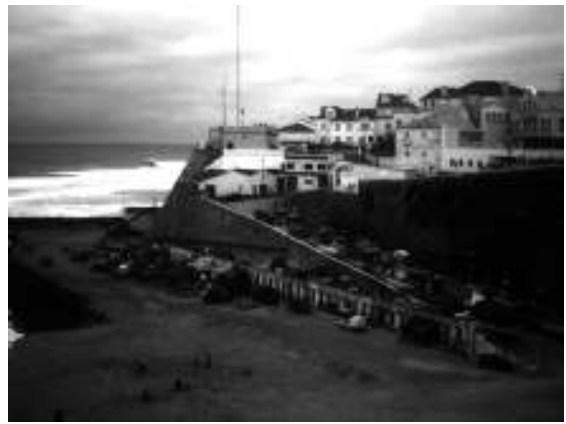

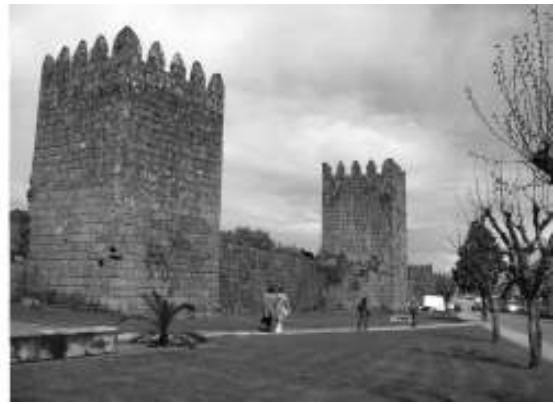

Figure 2: The first three pairs of source images, before printing. The images have been cropped, and one image in each pair has been horizontally flipped, to correspond to its position in the acquired images. Each image was then reduced in resolution and aligned to correspond, as well as possible, with the acquired images.

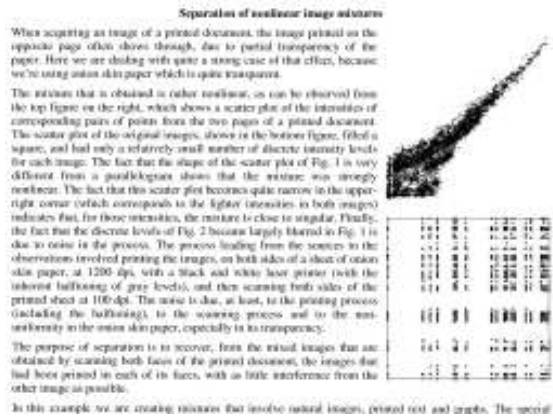

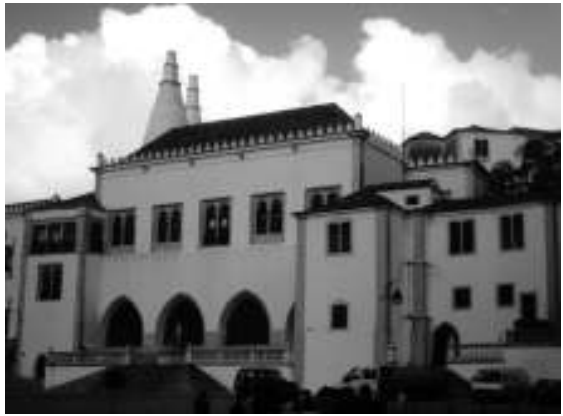
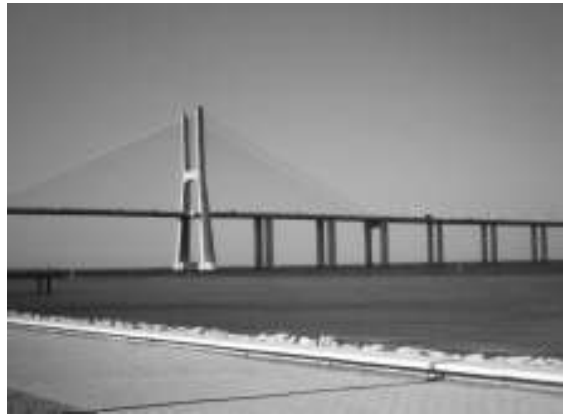

Figure 3: The fourth and fifth pairs of source images, before printing. One image in the last pair has been horizontally flipped. In the fourth pair no flipping has been performed, in order to keep the text's readability. Note, however, that the right-hand image of that pair appears flipped in the mixtures shown ahead.

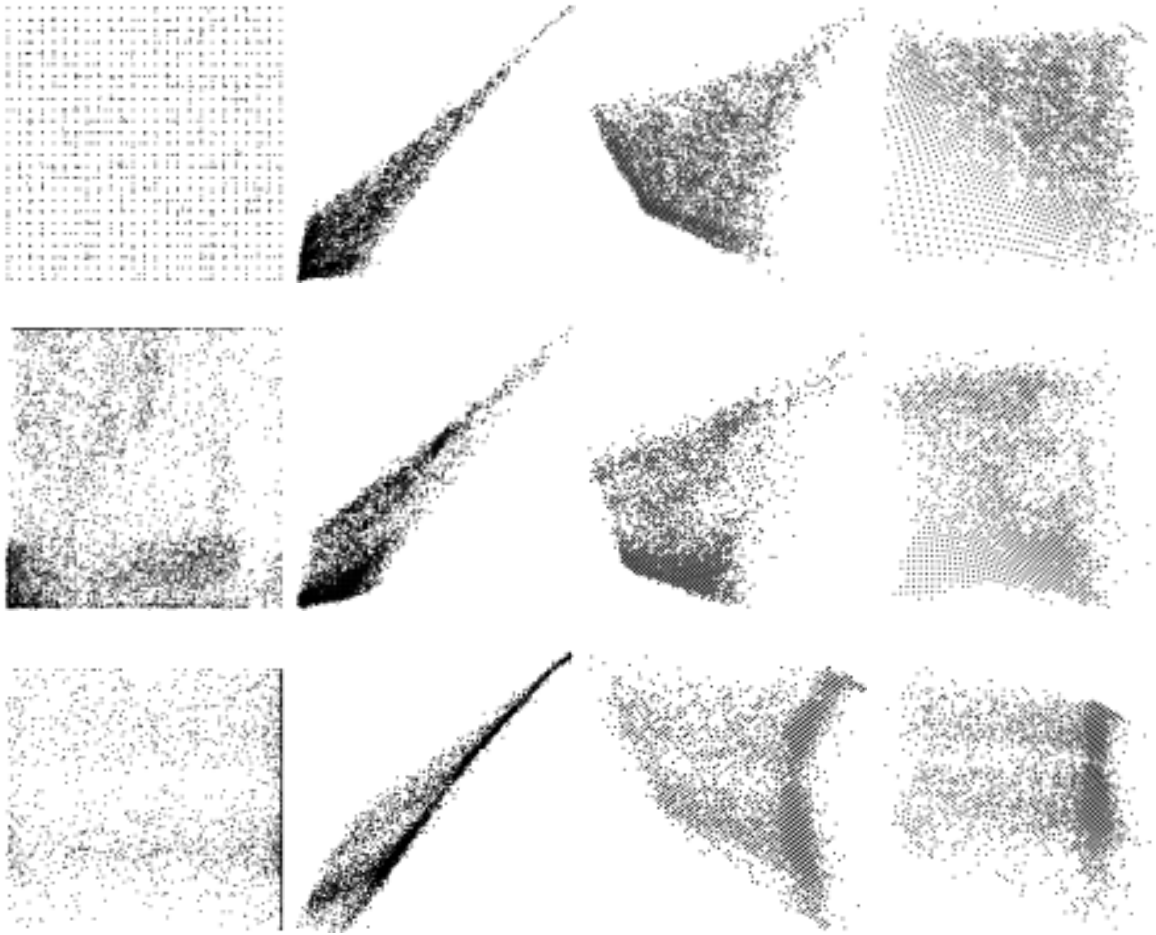

Figure 4: Scatter plots of the first three image pairs. From left to right: source images, acquired images, linear separation and nonlinear separation. The three rows correspond to the three pairs of images of Fig. 2. In each scatter plot, the horizontal axis corresponds to intensities from the left-hand image and the vertical axis to intensities from the right-hand image. The scale of each plot ranges from black (left/bottom) to white (right/top). Each scatter plot shows 5000 randomly selected points from the corresponding pair of images.

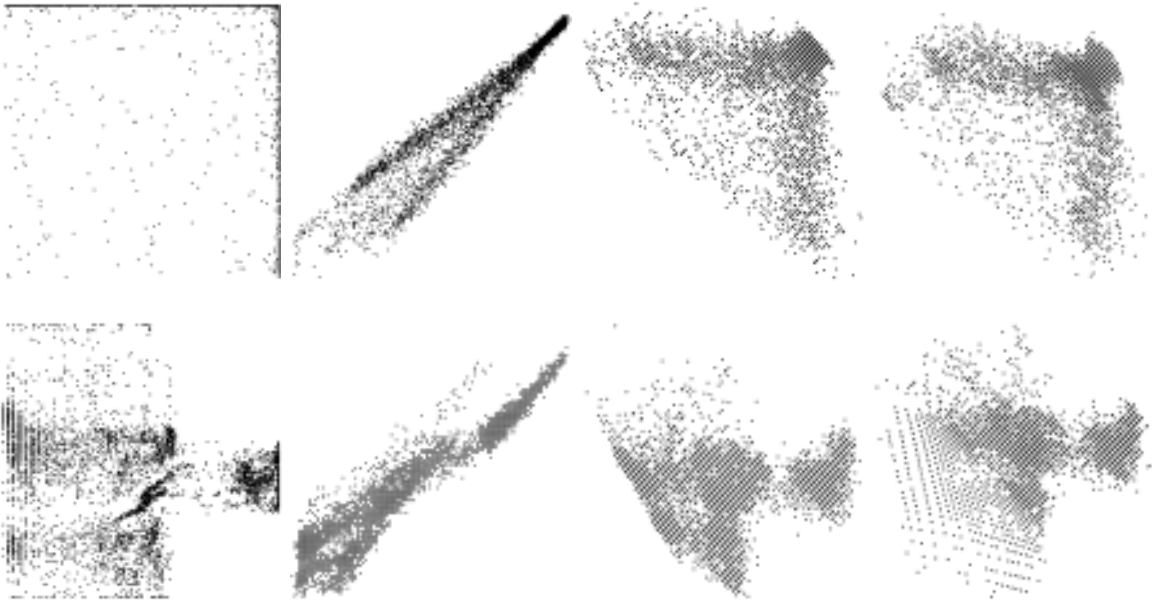

Figure 5: Scatter plots of the fourth and fifth image pairs. From left to right: source images, acquired images, linear separation and nonlinear separation. The two rows correspond to the two pairs of images of Fig. 3.

Some more comments are useful for a better understanding of the source distributions:

- The "grid" look of the first scatter plot reflects the fact that each of the source images had only 25 equally spaced intensities. Some intermediate intensities also appear in the plot due to the intensity interpolation performed in the resizing and alignment processes.

- The second scatter plot shows that, in this case, the two sources are almost independent from each other. The plot shows some evidence of saturation in the lightest intensities of the right-hand source image (vertical axis of the scatter plot). Since this saturation is in the source image, before printing, it should have no significant influence on the mixture and separation processes.

- The third and fourth scatter plots also show that the corresponding source pairs are approximately independent. The distributions of the sources that are images of text show that a very large percentage of their pixels is white. The non-white pixels show a continuous distribution, instead of just a black level, due to the interpolation performed in the resizing and alignment processes. The interpolation effect is much more noticeable here than in the first pair because, the character sizes being much smaller than the widths of the bars, many pixels fell on black-white edges, and only a very small percentage fell completely within black regions of the characters.

- The fifth scatter plot clearly shows that the sources of this pair are not independent. The plot shows some evidence of intensity quantization in the darkest levels of the left-hand source image (horizontal axis of the scatter plot), and of saturation in the lightest intensities of the same image. Since the quantization and saturation are in the source image, before printing, they should have no significant influence on the mixture and separation processes.

### 4.2 The mixture process: printing and acquisition

The images from each pair were printed on opposite faces of a sheet of onion skin paper. Printing was done with a 1200 dpi laser printer, using the printer's default halftoning system. Both faces of the sheet of onion skin paper were then scanned with a desktop scanner at a resolution of 100 dpi. This low resolution was chosen on purpose, so that the printer's halftoning grid would not be apparent in the scanned images. The scanner's "descreening" option (whose purpose is to minimize the visibility of the halftoning grid) was turned on.

We tried to keep the printing and acquisition processes as symmetrical as possible: the two source images in each pair were handled in an identical way, and the two acquired mixture images in each pair were also handled in an identical way. This implied disabling the scanner's "automatic image adjustment" feature, which adjusts the acquired image's brightness, contrast and gamma value in a manner that is not specified in the scanner's documentation.

The second column of scatter plots of Figs. 4 and 5 shows the joint distributions of the mixture components (after alignment, which is discussed in the next Section). The shapes of the mixture distributions show that the mixtures are nonlinear. This is especially clear in the first image pair, in which the joint distribution of the sources is approximately uniform within a square. A linear mixture process would have resulted in a mixture uniformly distributed within a parallelogram. The observed distribution has a shape that is far from a parallelogram and that is non-uniform, being more dense toward darker intensities than toward lighter ones. Both facts indicate that the mixture is nonlinear. The deviation from a parallelogram shape gives an idea of the amount of nonlinearity.

The mixture distribution, in the first pair, shows no traces of the discrete intensity levels that were present in the source images. This is due to noise introduced by the mixture process. This noise comes from three sources, at least: (1) the printing process, with the halftoning to reproduce grayscale levels; (2) the noise from the scanning process (from other tests of the same scanner this noise appears to be rather weak, essentially amounting to the intensity quantization into 256 levels), and (3) inhomogeneity of the onion skin paper (from our experience this appears to be the strongest source of noise). Later we'll have the possibility to have a better idea of the total amount of noise introduced by the mixture process.

On close inspection, the mixture scatter plots show that the points are arranged on a square grid. This is a result of the intensity quantization performed by the scanner.

### 4.3 Preprocessing

In the preprocessing stage, in each pair of acquired images one of them was first horizontally flipped, so that both images would have the same orientation. Then the images of each pair were aligned with each other by hand. In preliminary tests we found that even a very careful alignment, using translation, rotation and shear operations on the whole images, could not perform a good simultaneous alignment of all parts of the images. This was probably due to slight geometrical distortions introduced by the scanner. It indicated that an automatic, local alignment was needed. The use of the automatic local alignment relaxed the demands placed on the initial manual alignment.

In the alignment procedure that was finally adopted, the first step consisted just of a manual displacement of one of the images by an integer number of pixels in each direction, so that the two images would be coarsely aligned with each other. In a second step an automatic, local alignment was performed. For this, the resolution of both images was first increased by a factor of four in each direction, using bicubic interpolation. Then, one of the images was divided into $100 \times 100$ pixel squares (corresponding to $25 \times 25$ pixels in the original image), and for each square the best displacement was found, based on the maximum of the cross-correlation with the other image. The whole image was then rebuilt, based on these optimal displacements, and its resolution was reduced by a factor of 4. In this way a local alignment with a resolution of 1/4 pixel was achieved. Note that, although the alignment consisted only of local translations, it did handle the small rotations and shears that occur in problems of this kind, because these deformations consist just of different displacements for different points of the image. The fact that we used the same displacement for each $25 \times 25$ subimage caused only a negligible misalignment, relative to the true displacement that would be appropriate for each pixel.

There is a large variety of image alignment methods described in the literature, varying due to such aspects as the kinds of images to be aligned, the purpose of the alignment, etc. The reader can find an overview, somewhat oriented toward medical images, in Maintz (1998). The method that we used was designed specifically for handling the problem we needed to solve, but bears strong resemblances to some of the methods mentioned in that overview, and we make no claims to its originality.

As a final preprocessing step, the intensity range of each pair of images was normalized to the interval $[0, 1]$, 0 corresponding to the darkest pixel in the image pair and 1 to the lightest one. Figures 6 and 7 show the acquired images after preprocessing.[2]

As said above, we tried to keep the processing of both images in each pair as symmetrical as possible. An obvious asymmetry is due to the fact that only one image in each pair was modified in the alignment procedure. We used a high quality intensity interpolation method (bicubic) in the alignment procedure, so as to affect the image's quality as little as possible. The separation results that we present ahead, based on a symmetry constraint, seem to confirm that the mixture process was kept very close to symmetrical, despite the asymmetry in the alignment procedure.

---

2. All images of mixtures and of separation results displayed in this paper were adjusted in brightness and contrast so as to saturate the 1% brightest and 1% darkest pixels. This is a procedure that is commonly used for better display of images. This adjustment was performed for image display only: not for image separation and also not for the computation of quality measures.

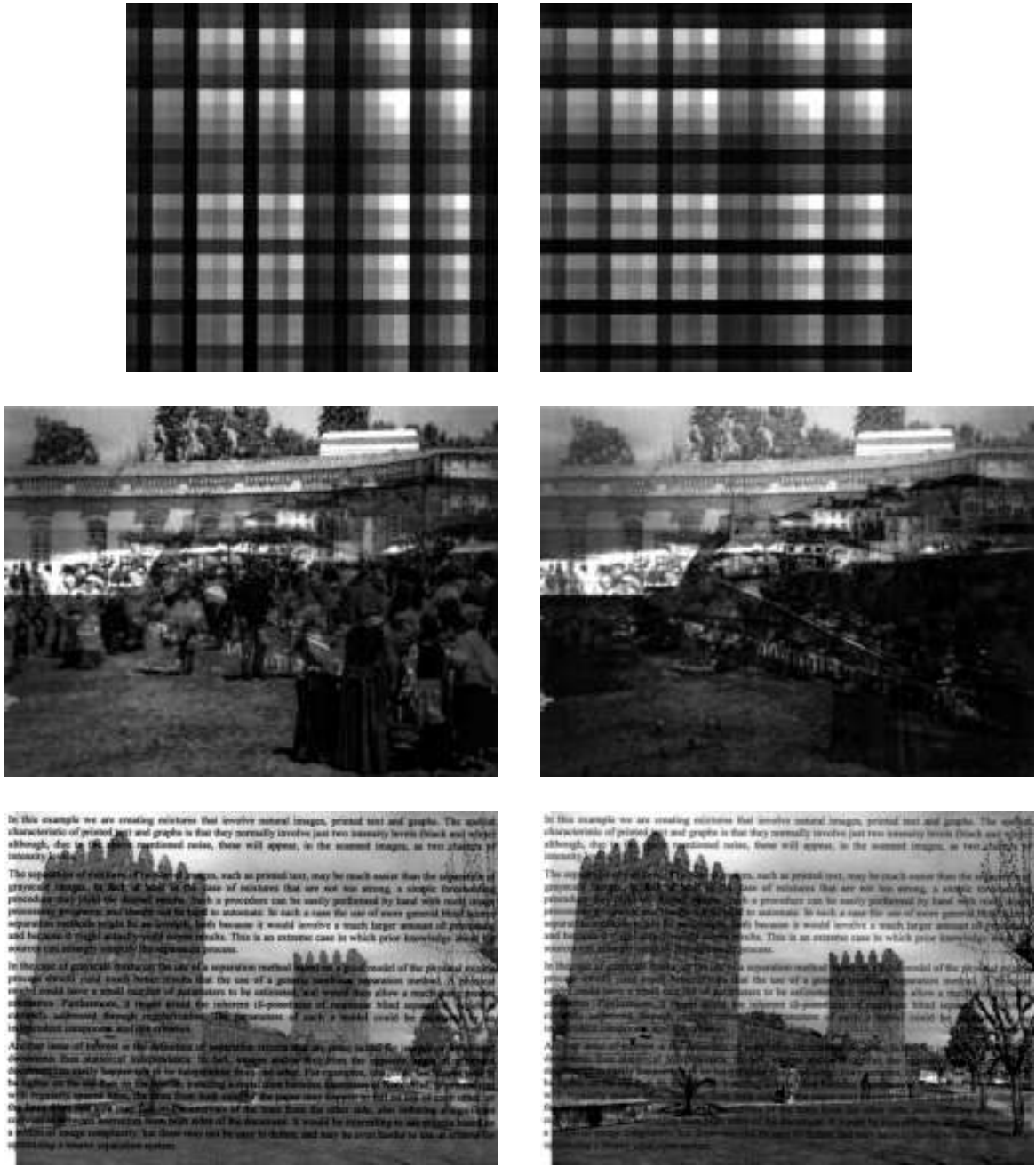

Figure 6: The first three pairs of acquired images, after preprocessing.

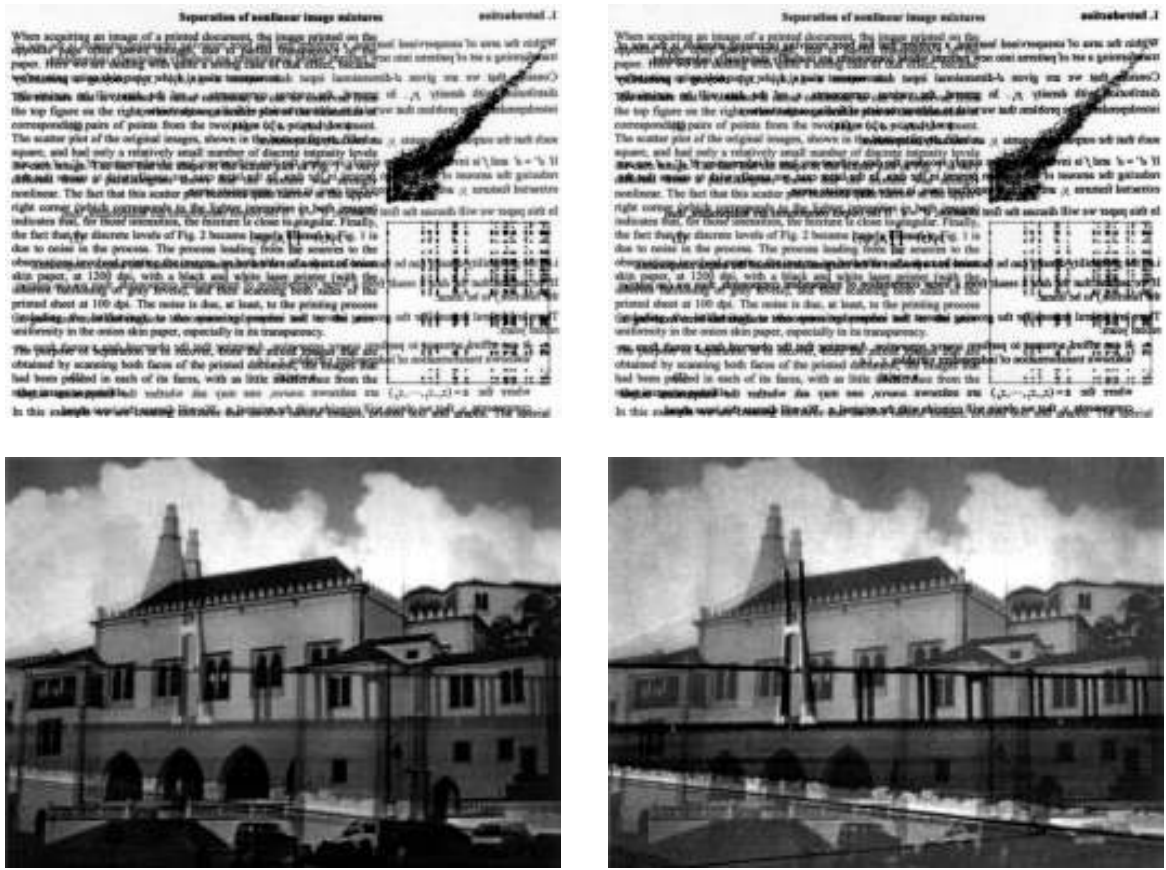

Figure 7: The fourth and fifth pairs of acquired images, after preprocessing.

## 5. Separation results

One of the main purposes of the work reported in this paper was to assess the viability and the advantage of performing nonlinear source separation, in a real-life nonlinear mixture problem, by means of an ICA-based separation system. Therefore we used source separation by linear ICA as a baseline for comparison. The next sections present the results of separation by linear and nonlinear ICA, followed by an assessment of the results with objective quality measures.

The mixture process that we used was as symmetrical as possible, so that an exchange of the source images should result just in a corresponding exchange of the mixture images (apart from noise). Therefore we applied symmetry constraints to the separation systems, as detailed ahead.

### 5.1 Linear separation

The linear ICA method that we used was MISEP with a linear $\mathbf{F}$ block, which corresponds to INFOMAX with adaptive nonlinearities. Each $\psi$ block was formed by an MLP with a single input and a single output, and with a hidden layer of 20 sigmoidal units. The output unit of each of these MLPs was linear, and there were no "shortcut" connections between input and output. The training set consisted of 5000 pairs of intensities, from randomly chosen pixel pairs of the acquired images. The $\mathbf{F}$ block was initialized with the identity matrix, and training was performed during 200 epochs, which were sufficient for convergence. The $\mathbf{F}$ block was constrained to be symmetrical. Symmetry was not enforced on the $\psi$ blocks because the distributions of the two sources were, in general, different from each other.

For each image pair, ten runs of the separation were made. These differed from one another in the selection of the 5000 pairs of pixels used to form the training set, and in the random initialization of the weights of the $\psi$ MLPs. The results of the ten runs were very similar to one another. Figures 8 and 9 show the results that were best, according to quality measure $Q_2$ (see Section 5.3). We see that a reasonable degree of separation was achieved in all cases, but some interference remained. The scatter plots in Figs. 4 and 5 (third column) show that, although a certain amount of separation was achieved, the nonlinear character of the mixture could not be undone by linear ICA, as expected. (Note: the arrangement of the scatter plots' points into lines – and, in fact, into a grid-like structure, although that is less apparent – is a result of the intensity quantization performed by the scanner).

### 5.2 Nonlinear separation

For nonlinear separation we used MISEP with a nonlinear $\mathbf{F}$ block. This block consisted of a multilayer perceptron with two inputs, two outputs and a hidden layer of 40 sigmoidal units. The output units were linear, and the hidden units were divided into two groups of 20, each group being connected to one of the output units. This MLP also had direct, "shortcut" connections between inputs and outputs. Since the output units were linear, the block could implement linear separation exactly, by setting the weights of the hidden layer's connections to zero.

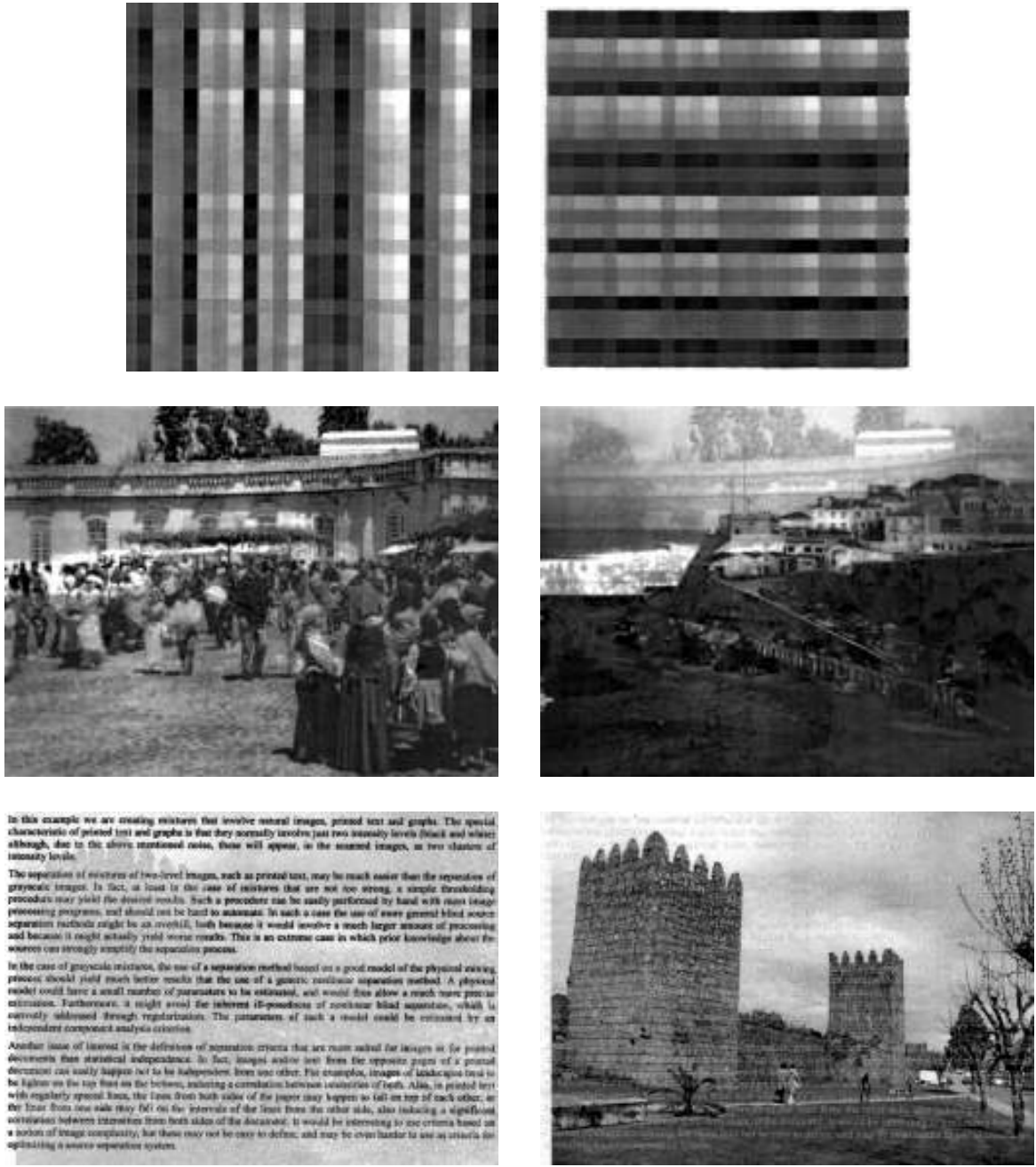

Figure 8: "Best" results of linear separation: first three image pairs.

Figure 9: "Best" results of linear separation: fourth and fifth image pairs.

As noted above, regularization plays an important role in dealing with the ill-posedness of nonlinear ICA. In our case regularization was achieved by three means: (i) initializing the $\mathbf{F}$ network to perform an identity mapping, (ii) constraining that network to be symmetrical, and (iii) constraining that network to be linear during the first 100 training epochs (by keeping the output weights of the hidden layer equal to zero during those epochs). Training was stopped at 400 epochs. At that point the progress of the optimization was in general very slow. As a test, in a few cases the optimization was extended to a much larger number of epochs, without any significant change in the separation results. Therefore the exact stopping point that was chosen doesn't appear to have had any significant influence on the results. The $\psi$ blocks had the same structure as in the linear separation case. Each 400-epoch training run took approximately 9 minutes on a 1.6 GHz Pentium-M (Centrino) processor.

For each image pair, ten runs of the separation were made, with different random selections of the 5000 pixel pairs forming the training set, and with different random initializations of the MLPs' weights (excluding, of course, those weights that were initially set to the identity matrix or to zero). Figures 10 and 11 show the best results that were obtained ("best" according to quality measure $Q_2$). The scatter plots corresponding to these separations are shown in the rightmost column of Figs. 4 and 5. Figures 12 and 13 show the worst separation results that were obtained ("worst" again according to $Q_2$).

## 5.3 Measures of separation quality

The images shown in the previous Section give an idea of the separation quality, but their evaluation is rather subjective. It depends on the viewer, as well as on other factors such as the conditions under which the images were printed or are viewed. Furthermore, a reasonable amount of image superposition can pass unnoticed in regions in which the "main" image has much variability. For these reasons we decided to also use objective measures of separation quality, which are not sensitive to such effects.

Experience with objective quality measures for nonlinear source separation is still very limited. This led us to compute four different quality measures. The first, that we denote by $Q_1$, was simply the signal to noise ratio (SNR) of the extracted component relative to the corresponding source.[3] We should note that, in a nonlinear separation context, the SNR, besides being sensitive to incomplete source separation and to noise, is also sensitive to any nonlinear transformation of the intensity scale that may be caused by the mixture and separation processes. It is well known that, in linear separation, the sources are recovered with unknown scale factors. In nonlinear ICA-based separation, each recovered source may be subject to an unknown nonlinear, invertible transformation. Measure $Q_1$ gives a global indication of the distortion of the extracted component relative to the corresponding source, including any nonlinear transformation of the intensity scale, besides including incomplete separation and noise.

Due to the possible presence of a nonlinear transformation of the intensity scale, our other three quality measures were defined so as to be invariant to such transformations. The second quality measure, $Q_2$, was a signal to noise ratio, modified so that it had the

---

3. For the computation of all quality measures we used the resized and aligned source images.

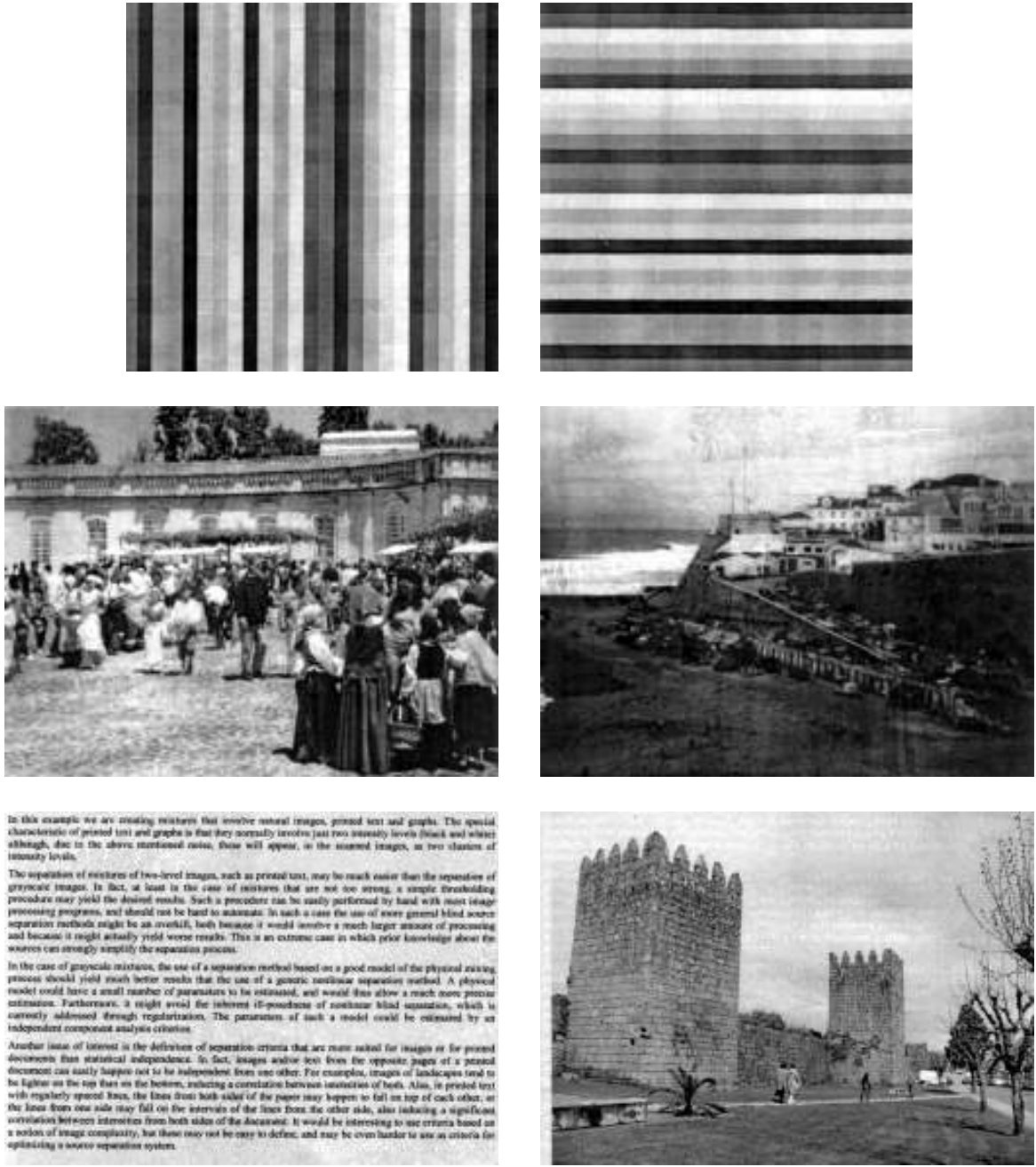

Figure 10: "Best" results of nonlinear separation: first three image pairs.

Figure 11: "Best" results of nonlinear separation: fourth and fifth image pairs.

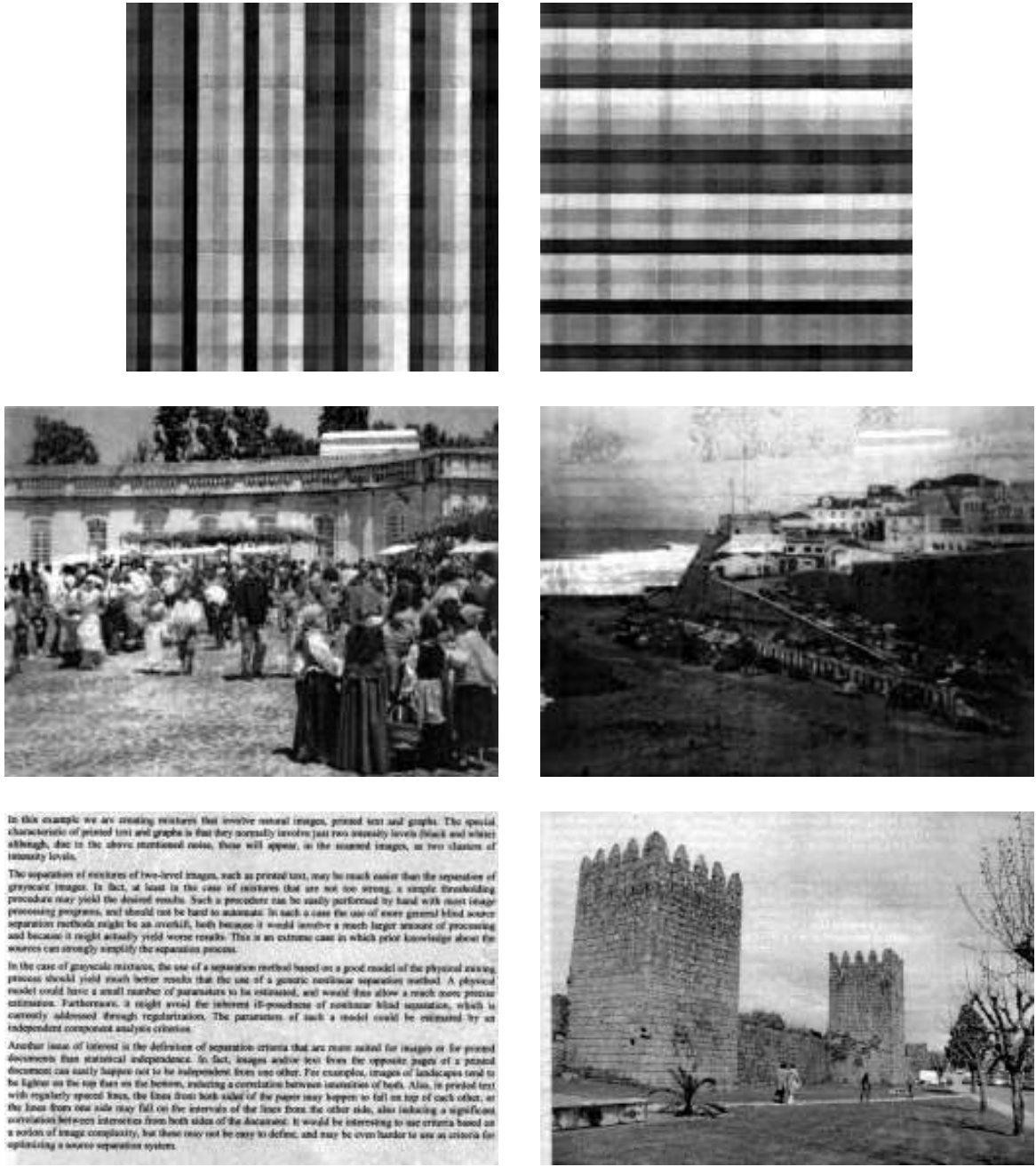

Figure 12: "Worst" results of nonlinear separation: first three image pairs.

Figure 13: "Worst" results of nonlinear separation: fourth and fifth image pairs.

invariance property mentioned above. It was given by

$$Q_2 = \frac{\text{variance of } S}{\text{variance of } N} \, , \tag{6}$$

where $S$ was the source image and $N$ was the noise that was present in the extracted component. This noise was computed as

$$N = f(Y) - S, \tag{7}$$

$Y$ being the extracted component, and $f$ being a nonlinear, monotonic transformation chosen so that $Q_2$ was maximal. In other terms, we chose a nonlinear, monotonic transformation of the intensity scale of the extracted component that made it become as close as possible to the corresponding source in SNR terms, and then used its SNR as the quality measure. The optimal $f(\cdot)$ was computed in table form. This was possible because the number of intensity levels in each image is finite, since each image has a finite number of pixels.

The other two measures that we used were information-theoretic:

- $Q_3$ was the mutual information between each extracted component and the corresponding source. The mutual information was estimated from a set of 5000 randomly selected pixel pairs, chosen independently from those forming the training set, and was computed using the $I^{(1)}$ estimator described in Kraskov et al. (2004), with $k = 3$ ($k$ is the nearest neighbor order used in that estimation algorithm; its recommended range, given in that reference, is between 2 and 4).

- $Q_4$ was the mutual information between each extracted component and the opposite source, computed in the same manner as for $Q_3$.

Note that other quality measures could easily be envisaged. For example, $Q_4 - Q_3$ would be a measure similar in spirit to the well known Amari index (Amari et al., 1996), but based on mutual information, to account for nonlinearities, and using a difference instead of a quotient due to its logarithmic character.

Another kind of measure that might come to mind would be similar to $Q_4$ (indicating the amount of interference, from the "wrong" source, that is present in the extracted component) but measured in terms of SNR instead of mutual information. Such a measure would not have made much sense, however, because in a nonlinear context the interference can be "positive" in some parts of the image and "negative" in other parts. These positive and negative parts would tend to cancel out. Therefore such a measure could sometimes indicate a misleadingly low amount of interference. In a measure like $Q_4$, based on mutual information, such positive and negative interferences do not cancel out, but instead have a cumulative effect.

As a reference for assessing the amount of separation achieved by the various methods, we show in Table 1 the values of the quality measures for the mixture components after preprocessing, without any separation.

The mean values of the quality measures for each of the ten-run series of separations are shown in Table 2. Note that for $Q_1$, $Q_2$ and $Q_3$ higher values are best, while for $Q_4$ lower values are best. The cases in which the difference between linear and nonlinear separation was significant at the 95% confidence level are shown in bold in the table.

| Image pair | Quality measure | No separation | |
| --- | --- | --- | --- |
| | | source 1 | source 2 |
| 1 | $Q_1$ | 1.9 | 1.9 |
| | $Q_2$ | 12.1 | 12.2 |
| | $Q_3$ | 1.21 | 1.23 |
| | $Q_4$ | 0.48 | 0.49 |
| 2 | $Q_1$ | -1.7 | 6.0 |
| | $Q_2$ | 8.7 | 12.3 |
| | $Q_3$ | 1.11 | 1.34 |
| | $Q_4$ | 0.56 | 0.60 |
| 3 | $Q_1$ | -4.5 | 6.6 |
| | $Q_2$ | 15.4 | 15.4 |
| | $Q_3$ | 0.38 | 1.65 |
| | $Q_4$ | 1.35 | 0.12 |
| 4 | $Q_1$ | 0.9 | -2.3 |
| | $Q_2$ | 17.1 | 16.7 |
| | $Q_3$ | 0.56 | 0.29 |
| | $Q_4$ | 0.23 | 0.43 |
| 5 | $Q_1$ | 9.6 | -6.4 |
| | $Q_2$ | 16.7 | 10.9 |
| | $Q_3$ | 1.85 | 1.07 |
| | $Q_4$ | 0.86 | 1.18 |

Table 1: Values of the objective quality measures for the unseparated mixture components. In this and in the following table $Q_1$ and $Q_2$ are given in dB and $Q_3$ and $Q_4$ in bits.

| Image pair | Quality measure | Linear separation | | Nonlinear separation | |
|:---:|:---:|:---:|:---:|:---:|:---:|
| | | source 1 | source 2 | source 1 | source 2 |
| 1 | $Q_1$ | 9.0 | 8.7 | **13.8** | **13.1** |
| | $Q_2$ | 17.8 | 17.6 | **20.6** | **20.2** |
| | $Q_3$ | 2.03 | 1.96 | **2.45** | **2.39** |
| | $Q_4$ | 0.48 | 0.46 | **0.23** | **0.26** |
| 2 | $Q_1$ | 5.2 | 10.5 | **9.3** | **13.9** |
| | $Q_2$ | 13.1 | 16.3 | **16.0** | **18.3** |
| | $Q_3$ | 1.56 | 1.78 | **1.83** | **1.95** |
| | $Q_4$ | 0.37 | 0.53 | **0.24** | **0.40** |
| 3 | $Q_1$ | 4.5 | 11.2 | **6.2** | 11.2 |
| | $Q_2$ | 19.3 | 19.7 | **20.6** | **20.9** |
| | $Q_3$ | 0.80 | 1.99 | **0.85** | **2.11** |
| | $Q_4$ | 0.36 | 0.18 | **0.09** | **0.15** |
| 4 | $Q_1$ | 5.8 | 3.4 | **6.0** | **3.7** |
| | $Q_2$ | 20.2 | 20.1 | **20.5** | **20.5** |
| | $Q_3$ | 0.74 | 0.48 | 0.75 | **0.51** |
| | $Q_4$ | 0.11 | 0.16 | 0.11 | 0.16 |
| 5 | $Q_1$ | 13.4 | **6.6** | **14.2** | 6.4 |
| | $Q_2$ | 19.7 | **19.0** | **20.3** | 18.9 |
| | $Q_3$ | 2.13 | **1.34** | **2.19** | 1.29 |
| | $Q_4$ | 0.71 | 0.46 | **0.56** | 0.49 |

Table 2: Objective quality results. The results shown are the average for each of the sets of ten test runs. The best result for each case is shown in bold when the difference (linear versus nonlinear) was significant at the 95% confidence level. For $Q_1$, $Q_2$ and $Q_3$ higher results are better, while for $Q_4$ lower results are better.

The measure that seemed to correlate best with our subjective evaluation of separation quality was $Q_2$, and this is why we chose it for the selection of the "best" and "worst" examples shown in Sections 5.1 and 5.2. The next best was $Q_1$. $Q_4$, which was intended to measure the amount of interference from the "wrong" source, was the one which correlated worst with our subjective quality evaluation.

## 5.4 Assessment of the results

For the first three image pairs, both the objective quality measures and our subjective evaluation showed a clear advantage of nonlinear separation over linear separation. Even the worst results of nonlinear separation seemed to be better, in general, than the best results of linear separation. Comparison of the third and fourth columns of scatter plots (Figs. 4 and 5) also confirms the advantage of nonlinear separation. This advantage was not so clear, however, for the fourth and fifth image pairs. We discuss now why we think this was so.

For the fourth image pair, most objective quality measures still show an advantage of nonlinear separation, but this advantage is very small, and our subjective evaluation showed the results of linear and nonlinear separation to be very similar in quality. This is also confirmed by comparing the corresponding scatter plots in Figs. 4 and 5. In this image pair, most pixels are white in at least one of the sources. The source scatter plot is dominated by two lines of points, located on the top and right-hand edges of the plot. This has the consequence that, with the specific mixture that was involved in the problem under study, linear ICA was able to perform a rather good separation. We see from the scatter plot of the linearly separated components that the lower-left area, corresponding to simultaneously dark pixels on both sources, was left unfilled by linear ICA. But this represented a rather small percentage of pixels, and had little impact on the overall separation quality.

We also see, from the rightmost scatter plot, that nonlinear separation also left the lower-left area unfilled. This may seem to be due to an incomplete optimization, but we tried extending the optimization to a much larger number of epochs without any significant change in the results. It is possible that the result shown corresponds to a local optimum. By playing with the network structure, with the initial conditions and with the constraints, we were sometimes able to get a result in which the lower left area of the scatter plot was filled. However, this made very little difference in the subjective or objective quality of the separation.

The results for the fifth image pair show that one of the sources was best separated by the linear method, while the other was best separated by the nonlinear one. But the differences between the two methods were rather small, even though most of them were statistically significant. Nonlinear separation apparently suffered a negative impact from the fact that the sources were not independent from each other and we were using independence as the separation criterion. The nonlinear separation network had many more degrees of freedom than the linear one, and used them to try to make the extracted components more independent from each other. In doing so it impaired the separation of one of the sources, instead of improving it, since the actual sources were not independent.

An important aspect of the results that we obtained is that, although the mixture process was nonlinear, and nonlinear separation could, in principle, introduce an arbitrary nonlinear transformation in each separated component, the total amount of nonlinearity introduced by the mixture and separation processes was relatively small. This is clear from the separation images that were shown (which were only normalized in brightness and contrast, as mentioned above) and from the values of the $Q_1$ measure. We also illustrate this, in a more clear form, in Fig. 14. This figure shows a scatter plot of the first extracted component versus the corresponding source, for the "average" case of the first image pair (the "average" case was chosen as the one whose value of $Q_2$ was closest to the average for the ten runs).

From our experience, there were two factors that were important in achieving this low level of nonlinearity. One was the fact that we linearly "primed" the separation network, by constraining it to be linear during the first 100 epochs. The other factor was that we gave a great amount of flexibility to the $\psi$ networks, by implementing them with a large number of hidden units. In previous tests in which these networks had just 6 hidden units, the separation results, as measured by $Q_2$, $Q_3$ or $Q_4$ were not very different from those presented here, but there often was a significant amount of nonlinearity introduced in the extracted components. This seems to have been caused by the $\mathbf{F}$ block trying to compensate for the limitations of the $\psi$ networks which could not, by themselves, make the distribution of each $Z_i$ close to uniform.

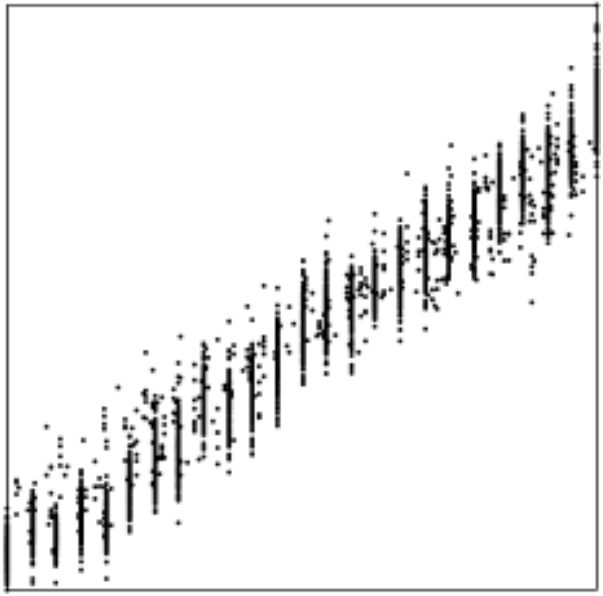

Figure 14: Scatter plot of the first extracted component versus the corresponding source, in an "average" run on the first image pair. Horizontal axis: source; vertical axis: extracted component.

There are some other aspects of the results, and of the experience that we gained in studying this problem, that are worth discussing. One of them has to do with the amount of noise introduced by the mixture process. We can take advantage of the fact that the source images that contain text have a large percentage of purely white pixels, which show up as strong, very thin lines in the corresponding scatter plots in the first column of Figs. 4 and 5, for having an idea of the amount of noise present in the mixtures and in the separated components. After the mixture, and also after linear or nonlinear separation, these lines appear broadened in the scatter plots, looking like fuzzy dark bands. The widths of these bands gives an idea of the amount of noise that was introduced by the mixing, or by the mixing plus separation. In the separation results the noise represents a significant percentage of the whole intensity range. Note that the separation process does not, by itself, introduce any noise. However, since it essentially consists of performing a weighted difference between the two mixture components, it does increase the amount of noise that is present, in relative terms.

Another interesting aspect has to do with understanding the "scale" of the quality measures based on mutual information (especially of $Q_3$ since, as we've already said, $Q_4$ seemed to be less meaningful). We were surprised by the relatively low values of mutual information between source and extracted component, even when the images looked well separated and $Q_2$ indicated relatively high SNR values after compensation of nonlinearities. For natural scene images, the mutual information between source and extracted component was roughly around 2 bits, while for text images it was below 1 bit. We can also observe from Table 2 that, for each source image, a change of 1 dB in SNR (i.e. in $Q_2$) corresponded, approximately, to a change of 0.1 bit in $Q_3$. Small changes in the value of mutual information seem to be much more significant than we expected before performing these tests.

An important aspect of the mixture process, that we have not mentioned so far, is that it didn't seem to be a purely point-wise process. The intensity of each source image at each point appeared to affect the observed mixture intensities in a small neighborhood of that point. This is especially noticeable by closely examining the separation results in the cases in which the image to be suppressed was a text image. The cause of this phenomenon probably was some lateral diffusion of light inside the paper. The effect was relatively weak at the scanning resolution that we used, but should become more pronounced at higher resolutions. A more perfect separation system should take this into account. However, non-point-wise nonlinear ICA is still essentially an unstudied topic, and is beyond the scope of this paper.

Another important aspect has to do with the use of the symmetry constraint. We were careful in ensuring that, both during scanning and in the preprocessing stage, both sides of the paper were handled in the same way. This allowed us to use a symmetry constraint in the separation networks. Such symmetry conditions in the mixture can probably be obtained when using a system like our desktop scanner, in which the paper has to be flipped, and the same set of sensors is used to acquire both sides. However, industrial scanners, which are used to digitize large quantities of documents, normally acquire both sides of the document at the same time, using two different sets of sensors. Such scanners often are strongly non-symmetric.[4] In such cases the symmetry constraint couldn't probably be used, or would have to be used only in an initial part of the training, after which it would have to be relaxed. We had no access to images from such scanners, and therefore couldn't assess what degree of separation would be achievable with them.

Still regarding a possible application to an actual scanning or photocopying device, there are two other aspects worth mentioning. One is that it doesn't seem to be possible to have a fixed separator, optimized at the factory for a specific device. This is because the mixture depends at least on the paper being used, and possibly also on the printing ink, halftoning process and other similar factors. It seems possible, however, to develop a physical model of the mixture process, with a small number of parameters, and then to find (algebraically or by approximate means) a parameterized inverse system. Its parameters may then be estimated through an ICA criterion. MISEP seems suited for this task, since it can use essentially any parameterized nonlinear system in the $\mathbf{F}$ block.

Another practical aspect has to do with the possible warping (existence of ripples) in the document being processed. We found that even very weak ripples, barely noticeable in the scanned images, would result in very strong light and dark bands in the separated images, both with linear and with nonlinear separation. This was, of course, a situation in which the mixture was spatially variant, and could not be adequately undone by a spatially invariant system. In our case we solved the problem by applying a very strong pressure to the cover of the scanner while scanning the documents, in order to eliminate the ripples. This might become an important issue in a practical application.

---

4. We acknowledge A. Shustorovitch for useful information regarding industrial scanners.

## 6. Conclusion

We showed an application of ICA to nonlinear source separation in a real-life problem of practical interest. One of the main issues that have been discussed in the last few years, concerning nonlinear ICA, is whether its inherent ill-posedness can be handled in practical situations. Our results show that it can, at least in this specific problem. We should say, however, that it took quite a bit of experimentation to find a set of conditions that could be used for all image pairs, yielding a good separation with relatively little variability in the separation results. In an earlier work (Almeida and Faria, 2004) we had not yet been able to achieve an adequate form of regularization, without resorting to an **F** block with a specialized form.

We presented comparisons of MISEP-based nonlinear ICA with linear ICA, one of the main purposes being to demonstrate the feasibility and the advantage of nonlinear source separation through ICA in a practical situation. It would also be very interesting to compare the nonlinear separation results presented here with those obtained with other nonlinear separation methods, such as ensemble learning (Lappalainen and Honkela, 2000), kernel-based nonlinear ICA (Harmeling et al., 2003) or geometric ICA (Theis et al., 2003). That comparison would have been outside the scope of the present paper. First of all, it would have involved a very large additional amount of work. Furthermore, the results obtained with a specific method are often much better if the method is tuned by someone experienced in its use. We have a reasonable amount of experience in using MISEP, but virtually no experience with any of the other methods. To enable comparisons we chose to make our test data, as well as our separation routines, available online (see the end of Section 4.3).

Future work will address several different issues, among which we can mention:

- The development of separation criteria that are more adequate for this problem than statistical independence. We have seen that, in this problem, the images to be separated may happen not to be independent. In such a case the quality of separation suffers. A more adequate separation criterion would not cause such degradation and might also be able to overcome much of the ill-posedness of nonlinear ICA, decreasing the dependence on regularization.

- The use of the spatial redundancy of images to reduce the ill-posedness of the problem, hopefully achieving separation with less dependence on regularization. Some published results (Harmeling et al., 2003) suggest that the use of signal structure may help to separate nonlinear mixtures with much reduced ill-posedness. That may make kernel-based nonlinear ICA a good candidate for handling this problem.

- The study of models of the mixture process that involve relatively few parameters. It seems possible to develop physically based and/or empirical models that depend on a few parameters (such as paper transparency and reflectivity, among others). Having few parameters, such models may have no ill-posedness, and may also be able to easily handle non-symmetrical systems.